\pdfoutput=1

\documentclass[11pt]{article}

\usepackage[preprint]{acl}

\usepackage{times}
\usepackage{latexsym}
\usepackage{booktabs} 
\usepackage{multirow} 
\usepackage{array}    
\usepackage{algorithm} 
\usepackage{algpseudocode}
\usepackage{amsmath}
\usepackage{tcolorbox}
\usepackage{enumitem}

\usepackage[T1]{fontenc}

\usepackage[utf8]{inputenc}

\usepackage{microtype}

\usepackage{inconsolata}

\usepackage{graphicx}

\usepackage{xcolor}

\title{An Empirical Study of LLM Reasoning Ability Under Strict Output Length Constraint}

\author{
 \textbf{Yi Sun\textsuperscript{1}},
 \textbf{Han Wang\textsuperscript{2}\thanks{\ Work was done while the authors were interning at Institute for AI Industry Research (AIR), Tsinghua University.}},
 \textbf{Jiaqiang Li\textsuperscript{3}\footnotemark[1]},
 \textbf{Jiacheng Liu\textsuperscript{4}\footnotemark[1]},
 \textbf{Xiangyu Li\textsuperscript{1}},
 \textbf{Hao Wen\textsuperscript{1}},
 \textbf{Yizhen Yuan\textsuperscript{1}},
 \\
 \textbf{Huiwen Zheng\textsuperscript{5}},
 \textbf{Yan Liang \textsuperscript{5}},
 \textbf{Yuanchun Li\textsuperscript{1}\thanks{\ Corresponding authors: Yuanchun Li and  Yunxin Liu.}},
 \textbf{Yunxin Liu\textsuperscript{1}\footnotemark[2]},
 \\
 \textsuperscript{1}Institute for AI Industry Research (AIR), Tsinghua University,\\
 \textsuperscript{2}Beijing University of Posts and Telecommunications,\\
 \textsuperscript{3}East China University of Science and Technology,\\
 \textsuperscript{4}Beijing Institute of Technology, 
 \textsuperscript{5}GDS Holdings Limited
}

\begin{document}
\maketitle

\begin{abstract}
Recent work has demonstrated the remarkable potential of Large Language Models (LLMs) in test-time scaling. By making models think before answering, they are able to achieve much higher accuracy with extra inference computation.
However, in many real-world scenarios, models are used under time constraints, where an answer should be given within a certain output length. It is unclear whether and how the reasoning ability of different LLMs remain effective under strict constraints.
We take a first look at this problem by conducting an in-depth empirical study. Specifically, we test 30 LLMs on common reasoning datasets under a wide range of output length budgets, and we analyze the correlation between the inference accuracy and various properties including model type, model size, prompt style, etc. We also consider the mappings between token budgets and actual on-device latency budgets.
The results have demonstrated several interesting findings regarding the budget-aware LLM reasoning ability that differ from the unconstrained situation, e.g. the optimal choices of either model size or prompt style change under different budgets. These findings offer timely evaluation to this area and practical guidance for users to deploy LLMs under real-world latency constraints.
\end{abstract}

\section{Introduction}
\definecolor{orange}{HTML}{FFB22C}
\definecolor{green}{HTML}{3A7D44}
\definecolor{purple}{HTML}{500073}






 





With the rapid advancement of Large Language Models (LLMs), there is a growing interest in their capabilities in tasks requiring advanced reasoning, such as programming, mathematical problem solving, and complex decision making. Their reasoning ability has become an important factor in the deployment of LLMs in real-world applications.


Various methods have been proposed to enhance the reasoning ability of LLMs. Some of them focus on equipping models with human-like cognitive processes and behaviors, such as Chain-of-Thought (CoT) reasoning \citep{1stcot}, self-correction \citep{kamoi-etal-2024-llms}, and multi-agent debating \citep{liang-etal-2024-encouraging, du2024improving} mechanisms.
Other approaches further enhance LLMs' reasoning by allocating more computational resources at test time to encourage deeper thinking, as seen in methods like self-consistency \citep{wang2023selfconsistency} and best-of-N decoding \citep{lightman2024lets}.
OpenAI o1 \citep{o1} and its open-source replicas, such as QwQ \citep{qwq-32b-preview} and Sky-T1 \citep{sky_t1_2025} exemplify the integration of these approaches,
using strategies like problem decomposition, multi-perspective reasoning, and error tracing to improve reasoning performance.

These methods boost LLMs' reasoning performance, but they also lead to lengthy reasoning steps, which incur considerable computation costs.
Recent works are beginning to explore strategies to optimize or control this cost, aiming to strike a balance between performance and reasoning efficiency \citep{han2024tokenbudgetawarellmreasoning, chen2024not, damani2024learninghardthinkinputadaptive, wang2025makepennycountdifficultyadaptive}. Approaches include dynamically adjusting the number of reasoning steps based on task difficulty \citep{manvi2024adaptiveinferencetimecomputellms, li2024escape}, imposing token limits in prompts to encourage concise responses \citep{han2024tokenbudgetawarellmreasoning}, and conducting token-aware training to incorporate length constraints at the embedding level \citep{takase-okazaki-2019-positional, butcher2024precise}.

However, prior research neglects scenarios in which LLMs' reasoning may be constrained by output limits. We believe this is an important setting that deserves more attention.
First, many real-world AI applications are time constrained, requiring rapid or even real-time decisions. For example, autonomous driving systems should make precise action predictions within a limited time frame \citep{wang2024drivecotintegratingchainofthoughtreasoning}. 
Second, time-constrained reasoning under deadlines is an important trait of human intelligence.
It is interesting to study whether and how LLMs preserve their reasoning ability under strict output length constraints.

Therefore, we conduct an empirical study of open-source LLMs' reasoning ability with \textbf{strict} output length constraints. Specifically, we test different models on various math datasets, while limiting the number of output tokens. This ensures that models' inference can be guaranteed to complete within time budgets, and achieved accuracies can be regarded as the actual performance of models in time-constrained settings.
We also analyze whether and how different factors (model type, model size and prompt style) can affect such performance.

To evaluate LLMs' reasoning under constrained scenarios, the naive approach is to directly terminate the generation process at the maximum token budget. However, this approach may lead to poor performance that does not reflect models' real capability,
because LLMs may not explicitly output final answers during the reasoning.
Instead, we adopt a more reasonable scheme named early stopping, where reasoning is interrupted at several tokens before budgets. A termination message \textit{"Time's Up! Therefore, the final answer is:"} is appended to the end of model output, inducing LLMs to generate final answers in a structured format during the continued inference, until the generated tokens reaching the total budget. By avoiding abrupt reasoning truncation, this scheme faithfully reflects LLMs' time-constrained reasoning capabilities. We will elaborate on these two methods in Section \ref{sec:method}.

Our experiments lead to several interesting and even surprising findings in Section \ref{sec:findings}. For example, when testing LLMs under output constraints, we observe the disagreement with scaling law \citep{scaling_law} and the change in golden models and prompts in different deployment settings. Carefully tailored reasoning models are also not necessarily better than normal instruction tuned models. When mapping token budget to latency budget on real devices, we find that medium sized models often achieve the best efficiency under strict latency limits, while larger models gradually surpass them as latency constraints are relaxed. 
We expect these findings to present a general impression of how existing LLMs perform under strict output length constraints, and give practitioners some useful guidance to deploy LLMs in time-sensitive scenarios.

Our contributions are as follows:

\begin{enumerate}[itemsep=0pt, topsep=0pt]
    \item We study an important scenario of LLM reasoning. To the best of our knowledge, this is the first thorough empirical study of LLM reasoning under strict output length constraints. 
    \item We conduct extensive experiments with a wide range of LLMs of various sizes and types. We evaluate their reasoning ability across mathematical datasets of varying difficulty.
    \item We summarize several interesting findings, which may be helpful for researchers to understand the time-constrained reasoning ability of LLMs and improve them accordingly, and for developers to make informed choices based on their deployment scenarios.
\end{enumerate}

Our code and data will be fully open-sourced.

\section{Related Work}



\textbf{Test-Time Scaling}. 
Rather than expanding model parameters and training data~\cite{scaling_law}, recent studies now emphasize test-time scaling to improve LLMs' reasoning capabilities ~\cite{o1}. 
This can be achieved by methods like repeated sampling~\cite{snell2024scalingllmtesttimecompute, ll_monkey}, sequential sampling~\cite{lee2025evolving, hou2025advancing}, and tree-based search~\cite{mcts_llm, chen2024alphamath, yao2023tree}. 
Furthermore, researchers began to explore training LLMs using reinforcement learning to think deeper and generate longer CoTs~\cite{o1, deepseekai2025deepseekr1incentivizingreasoningcapability}. 
Despite of improvement of these methods, LLMs' reasoning ability under strict output length constraints is still unexplored.



\textbf{Efficient Reasoning Techniques.}
Several works \citep{nayab2025concisethoughtsimpactoutput, han2024tokenbudgetawarellmreasoning} showcase that 
adding output length limits into prompts can encourage LLMs to generate more concise yet still correct responses.
Other works \citep{damani2024learninghardthinkinputadaptive, wang2025makepennycountdifficultyadaptive} allocate additional computation budget based on predicted complexity of queries, either by routing to larger models or conducting more samplings to vote for the final answer. 
Besides, other methods \citep{manvi2024adaptiveinferencetimecomputellms, li2024escape} allow for mid-generation control during multiple samplings to prune unpromising traces.
The focus of our work is not to present an optimized method to improve LLMs' reasoning efficiency. Instead, our contribution is the rational and elaborate evaluation of LLMs reasoning ability
under strict token budgets. Since there are evidence \citep{yuan2024followinglengthconstraintsinstructions, xu2025chaindraftthinkingfaster} that directly adding length limits into prompts for LLMs to follow often fails or even brings performance degeneration, we do not adopt this method to impose strict output length constraints for LLMs' reasoing.


\section{Method}
\label{sec:method}

\begin{figure}[htp]
    \centering
    \includegraphics[width=\linewidth]{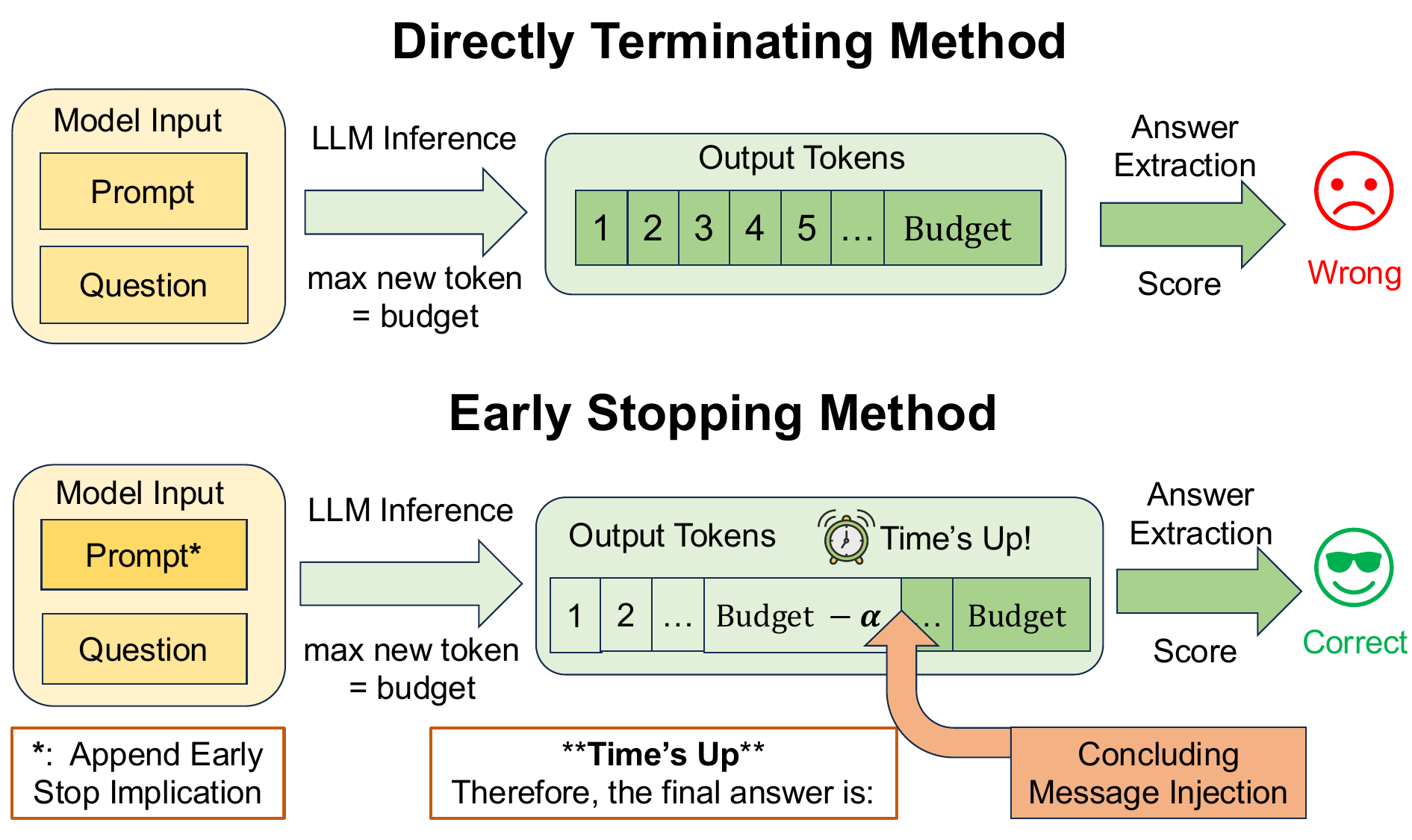}
    \caption{Two methods used in our work to ensure strict output length constraints for LLM reasoning.}
    \label{fig:method}
\end{figure}

We use two methods to impose output length constraints on LLMs' reasoning, as shown in Figure \ref{fig:method}.
They are referred to as \textbf{directly terminating} and \textbf{early stopping}.
Both methods can strictly ensure the generated tokens do not exceed given budgets.

\subsection{Directly Terminating at Token Budget}

The first method is to directly terminate LLMs' reasoning at token budgets. This is done by setting the parameter \textit{max\_new\_token} in LLM inference APIs, such as vLLM \citep{vllm}, to token budgets. 
The final answer is then extracted from all generated tokens. Details of the extraction and scoring procedures are provided in Appendix \ref{sec:appendix answer extraction and scoring}

This method is very straightforward but prone to underestimation of LLMs' reasoning ability under constrained scenarios, because abrupt truncation at token budgets during the reasoning process may result in incomplete responses. Only the problems whose responses fall within token budgets can be correctly solved. As token budgets increase, more problems will be solvable within the budgets, and the accuracy across the dataset will also improve.



\subsection{Early Stopping Before Token Budget}

To obtain more reliable understanding of LLMs' reasoning capability under output constraints, we propose to early stop reasoning before token budgets and instruct LLMs to conclude final answers at once. As shown in Figure \ref{fig:method}, this needs two modifications to directly terminating: appending early stop implication in model input construction and early stopping with concluding message injection. 

Before inference, we append the model prompt with some words to inform LLMs of the potential early stopping during the reasoning process. The keyword \verb|**Time's Up!**| is used as the signal of early stopping. 
During LLMs' reasoning, when the number of generated tokens reaches \(Budget - \alpha \), we will append concluding message with signal \verb|**Time's Up!**| in it, at the end of model outputs. 
Then LLMs are allowed to conclude their answers within \(\alpha\)\footnote{We set \(\alpha=25\) in all our experiments, which is large enough to cover the correct final answers of tested problems.} tokens, thus ensuring the total output length is still within token budgets. The final answer will be extracted and scored from tokens generated after concluding message injection. To ensure fairness, the procedures used here are the same as directly terminating method. If LLMs can finish their output generation using less than \(Budget - \alpha \) tokens, then concluding message injection and further inference are not needed. And the final answer will be derived from all output tokens, just like the directly terminating method.

We list the full version of token budget implication and concluding message in Appendix \ref{sec:appendix implication and concluding}. 

\section{Experiment Setup}


\subsection{Datasets}
To evaluate LLM's reasoning ability, we use the test splits of two math datasets: GSM8K \citep{GSM8k} and MATH500 \citep{MATH500}. 

\textbf{GSM8K} includes 8.5k grade school level math word questions with high linguistic diversity, designed to test LLMs’ ability to perform step-by-step reasoning. Its test split consists of 1319 problems.

\textbf{MATH500} is a scale extraction from the original MATH \citep{MATH} dataset, with high school to early college level math problems. It includes subtopics like algebra, geometry and number theory, designed to evaluate advanced mathematical reasoning and problem-solving skills of LLMs. There are 500 problems in total for testing.

\subsection{Models}
To enrich the evaluation within our work, we selected three types of open-sourced LLMs:

\textbf{Instruction models} include series like Qwen-2.5-Instruct \citep{qwen2.5,qwen2}, Phi-3-128k-instruct \citep{abdin2024phi3technicalreporthighly}, gemma-2-it  \citep{gemma_2024}, and Llama-3.2~\citep{meta2024llama32}. Other models include Llama-3.1-8B-Instruct \citep{grattafiori2024llama3herdmodels} , Ministral-8B-Instruct-2410 \citep{mistral2024ministraux}, Mistral-Nemo-Instruct-2407 \citep{mistral2024nemo}, Mistral-Small-Instruct-2409 \citep{mistral2024small} and Phi-4 \citep{abdin2024phi4technicalreport}.

\textbf{Math models} include those specially trained or fine tuned using mathematical data. We tested Qwen2.5-Math-Instruct \citep{yang2024qwen25mathtechnicalreportmathematical} and Mathstral-7B \citep{Mathstral}.

\textbf{Reasoning models} include QwQ-32B-Preview \citep{qwq-32b-preview, qwen2}, Sky-T1 \citep{sky_t1_2025}, DeepSeek-R1-Distill-Qwen series \citep{deepseekai2025deepseekr1incentivizingreasoningcapability}. They tend to generate longer reasoning steps than instruction or math models to scale up their problem solving ability.

\setlength{\textfloatsep}{10pt} 
\begin{table}[!tbp]
    \centering
    \small
    \setlength{\tabcolsep}{2.9pt}
    \begin{tabular}{ccccc}
        \toprule
        \textbf{Model} & \textbf{Size} & \textbf{GSM8K} & \textbf{MATH500} \\
        \midrule
        DRD-Qwen & 1.5B & 75.7(75.7) & 68.2(71.4) $\uparrow$&\\
        DRD-Qwen & 7B & 87.9(87.9) & 77.0(81.2) $\uparrow$&\\
        DRD-Qwen & 14B & 92.0(92.0) & 78.6(87.4) $\uparrow$&\\
        DRD-Qwen & 32B & 94.5(94.5) & 79.8(86.0) $\uparrow$&\\
        QwQ & 32B & 95.5(95.5) & 82.0(87.6) $\uparrow$\\
        Sky-T1 & 32B & 96.4(96.4) & 87.6(87.6) \\
        \midrule
        Qwen2.5-Math & 1.5B & 85.0 & 74.0 \\
        Qwen2.5-Math & 7B & 95.5 & 82.4 &\\
        Mathstral & 7B & 83.6 & 51.2 \\
        \midrule
        Qwen-2.5 & 1.5B & 73.9 & 53.0 &\\
        Qwen-2.5 & 3B & 85.7 & 65.8 &\\
        Qwen-2.5 & 7B & 91.9 & 75.6 &\\
        Qwen-2.5 & 14B & 94.8 & 79.0 &\\
        Qwen-2.5 & 32B & 95.9 & 81.0 &\\
        Qwen-2.5 & 72B & 95.8 & 83.2 &\\
        gemma-2 & 2B & 64.8 & 23.8 \\
        gemma-2 & 9B & 87.7 & 48.0 \\
        gemma-2 & 27B & 90.8 & 56.2 \\
        Llama-3.2 & 1B & 48.8 & 26.6 \\
        Llama-3.2 & 3B & 76.2 & 48.6 \\
        Llama-3.1 & 8B & 84.1 & 46.2 \\
        Llama-3.1 & 70B & 94.9 & 62.6 \\
        Ministral & 8B & 87.1 & 56.8 \\
        Mistral-Nemo & 12B & 85.6 & 43.6 \\
        Mistral-Small & 22B & 91.7 & 61.2 \\
        Phi-3-mini & 3.8B & 86.7 & 39.0 \\
        Phi-3-small & 7B & 88.9 & 50.8 \\
        Phi-3-medium & 14B & 88.0 & 49.6 \\
        Phi-3.5-mini & 3.8B & 86.8 & 45.2 \\
        Phi-4 & 14B & 95.1 & 79.2 \\
        \bottomrule
    \end{tabular}
    \caption{Accuracy of tested models on both datasets. LLMs are prompted with step by step style. Max new token for non reasoning models: 4096, for reasoning models: 4096 (8192). The $\uparrow$ sign means reasoning models' score can still increase if more tokens are allowed.}
    \label{tab:all_models}
\end{table}

\subsection{Prompts}
\label{sec:prompt design}
We use the following prompt styles to guide LLMs' reasoning in three different patterns:


\begin{enumerate}[itemsep=0pt, topsep=0pt]
    \item \textbf{step-by-step (sbs).} This is the most common style to elicit model reasoning ability.
    \item \textbf{coarse-to-fine (c2f).} This requests LLMs to give a coarse-grained reasoning summary before starting fine-grained reasoning steps. 
    
    \item \textbf{answer-and-verify (aav).} This style lets the models to give an initial answer quickly, then verify and revise it iteratively.
\end{enumerate}

We expect c2f and aav styles
can be used to better conclude final answers for LLMs if their reasoning is early stopped due to limited token budgets. 
We show the influence of prompt styles on LLMs' reasoning performance in Sec \ref{sec:findings 2}. 
The full version of prompt styles can be found in Appendix \ref{sec:appendix prompt style}.
In order to align with the LLMs' training procedure, we construct model inputs based on their default chat templates and guidance from Hugging Face \citep{wolf-etal-2020-transformers}. Model input construction process is also well illustrated in Appendix \ref{sec:appendix model input}.

\subsection{Evaluation Framework}
\label{sec:evaluation framework}

We use the evaluation framework from Qwen-2.5-Math \citep{yang2024qwen25mathtechnicalreportmathematical}, which supports our datasets and models. Zero-shot and greedy decoding strategy is used to guarantee performance consistency. Package version used in experiments is   transformers \citep{wolf-etal-2020-transformers} 4.46.3 and vLLM \citep{vllm} 0.6.3.post1.
To give an outline of models' basic reasoning ability on the evaluation framework, and simplify the reference to their names in following sections, we report their performance on both datasets in Table \ref{tab:all_models}.

\section{Results and Findings}
\label{sec:findings}
\definecolor{orange}{HTML}{FFB22C}
\definecolor{green}{HTML}{3A7D44}
\definecolor{purple}{HTML}{500073}

Through comprehensive evaluation and analysis of experiment results, we conclude five interesting findings, which lead to future directions worthy of exploration and some practical guidance for deploying LLMs under strict output length constraints. We also build a website\footnote{Examples can be found in https://time-is-up.github.io/} to show some examples, allowing readers to gain a clearer understanding when reading the following findings.

\begin{figure*}[!htbp]
    \includegraphics[width=2\columnwidth, trim=15 25 25 180, clip]{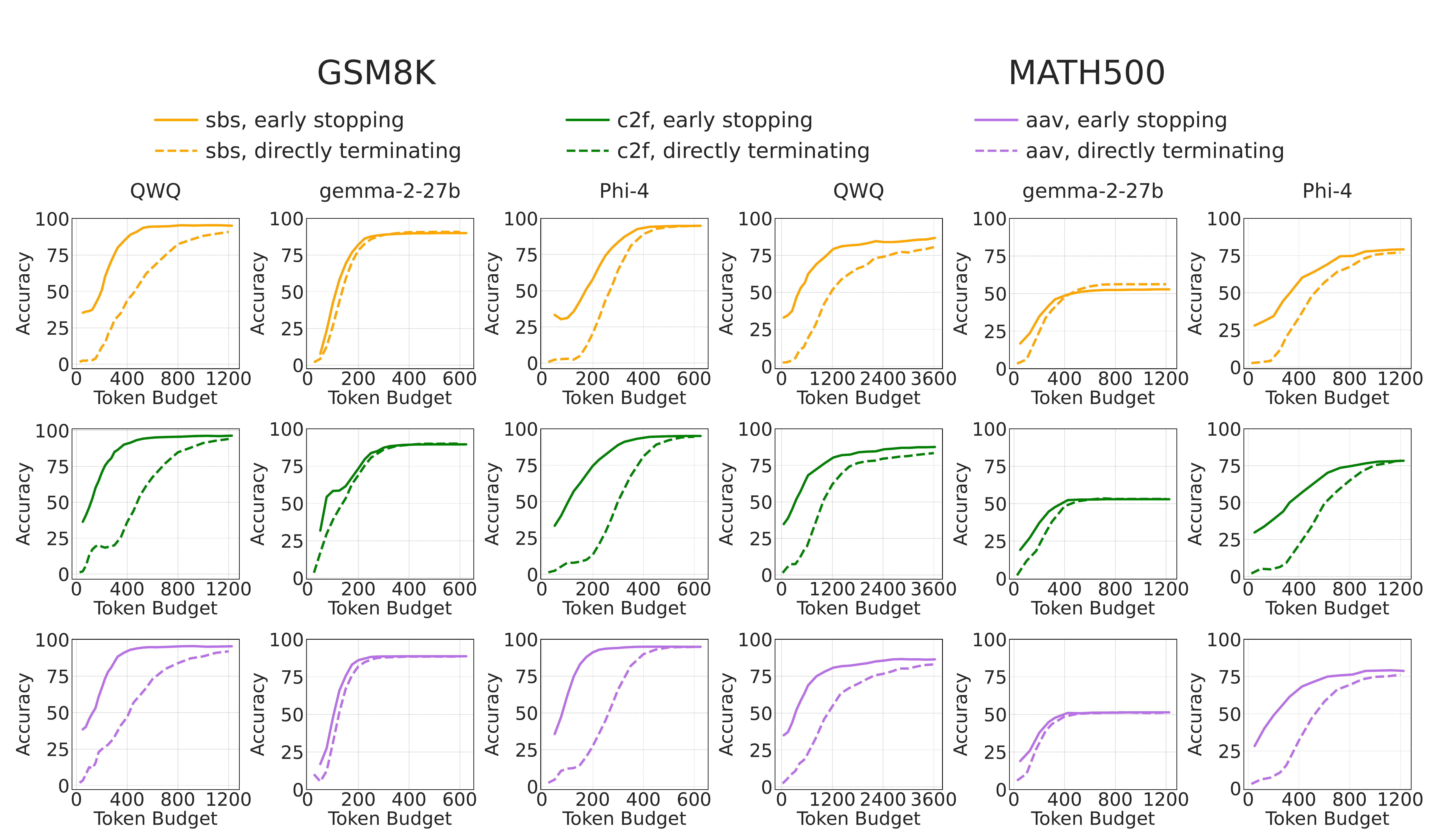}
    \caption {Early stopping method (solid line) outperforms directly terminating (dashed line) on GSM8K (left) and MATH500 (right) datasets. Prompting style: \textcolor{orange}{sbs}, \textcolor{green}{c2f} and \textcolor{purple}{aav}.}
    \label{fig:findings1_plot}
\end{figure*}

\begin{table*}[tbp]
    \centering
    \small
    \setlength{\tabcolsep}{4pt}
    \resizebox{0.95\textwidth}{!}{ 
    \begin{tabular}{l|p{0.75cm}p{0.75cm}p{0.75cm}|p{0.75cm}p{0.75cm}p{0.75cm}|p{0.75cm}p{0.75cm}p{0.75cm}|p{0.75cm}p{0.75cm}p{0.75cm}}
    \toprule
    \multirow{3}{*}{Model} & \multicolumn{6}{c|}{GSM8K} & \multicolumn{6}{c}{MATH500} \\
    \cmidrule(lr){2-7} \cmidrule(lr){8-13}
     & \multicolumn{3}{c|}{Budget 50} & \multicolumn{3}{c|}{Budget 150} & \multicolumn{3}{c|}{Budget 100} & \multicolumn{3}{c}{Budget 200} \\
    \cmidrule(lr){2-4} \cmidrule(lr){5-7} \cmidrule(lr){8-10} \cmidrule(lr){11-13}
     & \multicolumn{1}{c}{aav} & \multicolumn{1}{c}{c2f} & \multicolumn{1}{c|}{sbs} & \multicolumn{1}{c}{aav} & \multicolumn{1}{c}{c2f} & \multicolumn{1}{c|}{sbs} & \multicolumn{1}{c}{aav} & \multicolumn{1}{c}{c2f} & \multicolumn{1}{c|}{sbs} & \multicolumn{1}{c}{aav} & \multicolumn{1}{c}{c2f} & \multicolumn{1}{c}{sbs} \\
    \midrule
    Sky-T1 & \textbf{46.6} & 44.6 & 38.4 & \textbf{83.7} & 70.0 & 58.5 & 40.0 & \textbf{40.4} & 36.6 & \textbf{53.6} & 49.8 & 43.0 \\
    QwQ & 40.3 & \textbf{41.0} & 36.2 & 60.9 & \textbf{64.9} & 45.7 & 37.0 & \textbf{38.6} & 34.8 & 42.6 & \textbf{44.6} & 35.6 \\
    DRD-Qwen-1.5B & 9.7 & \textbf{11.8} & 9.9 & \textbf{48.6} & 48.4 & 45.8 & \textbf{19.0} & 17.0 & 18.2 & 32.4 & \textbf{33.0} & 32.2 \\
    DRD-Qwen-7B & 27.4 & \textbf{28.0} & 24.8 & \textbf{67.4} & 49.2 & 65.1 & \textbf{29.6} & 28.2 & 29.4 & 36.8 & 28.8 & \textbf{40.4} \\
    DRD-Qwen-14B & 27.5 & \textbf{29.1} & 27.7 & \textbf{68.2} & 66.0 & 62.9 & \textbf{32.4} & 29.4 & 30.8 & \textbf{41.8} & 40.8 & 34.4 \\
    DRD-Qwen-32B & 35.7 & \textbf{36.2} & 32.6 & \textbf{76.7} & 72.9 & 72.3 & \textbf{38.4} & 35.8 & 36.6 & \textbf{47.4} & 43.4 & 46.0 \\
    \midrule
    Qwen2.5-Math-1.5B & \textbf{16.7} & 15.0 & 15.9 & \textbf{31.9} & 31.1 & 31.5 & \textbf{24.6} & 22.4 & 22.0 & \textbf{31.6} & 31.4 & 31.5 \\
    Qwen2.5-Math-7B & 29.4 & \textbf{30.9} & 29.6 & \textbf{49.1} & 47.7 & 46.7 & \textbf{37.0} & \textbf{37.0} & 36.4 & \textbf{43.8} & \textbf{43.8} & 42.0 \\
    Mathstral-7B & \textbf{19.7} & 16.8 & 8.6 & 54.3 & \textbf{63.5} & 47.8 & \textbf{22.6} & 21.8 & 16.8 & \textbf{35.4} & 33.0 & 31.0 \\
    \midrule
    Qwen-2.5-1.5B & \textbf{21.2} & 19.8 & 9.8 & 43.4 & \textbf{49.6} & 25.7 & 18.2 & \textbf{19.0} & 15.4 & \textbf{27.0} & 26.8 & 25.2 \\
    Qwen-2.5-3B & 13.4 & \textbf{21.5} & 12.4 & 35.0 & \textbf{49.7} & 36.0 & 22.0 & \textbf{23.6} & 21.0 & 31.8 & \textbf{32.0} & 29.6 \\
    Qwen-2.5-7B & \textbf{46.4} & 35.0 & 20.5 & \textbf{78.2} & 60.1 & 48.1 & \textbf{39.2} & 31.6 & 26.6 & \textbf{50.0} & 43.4 & 35.4 \\
    Qwen-2.5-14B & \textbf{46.3} & 40.4 & 24.3 & \textbf{82.8} & 65.4 & 39.0 & \textbf{44.0} & 39.2 & 25.2 & \textbf{55.8} & 50.4 & 38.0 \\
    Qwen-2.5-32B & \textbf{57.6} & 47.8 & 36.2 & \textbf{88.8} & 76.1 & 56.3 & \textbf{50.2} & 44.0 & 38.4 & \textbf{60.4} & 53.2 & 46.2 \\
    Qwen-2.5-72B & \textbf{52.0} & 43.4 & 37.0 & \textbf{85.6} & 75.6 & 57.8 & \textbf{48.4} & 45.0 & 44.0 & \textbf{58.0} & 54.0 & 48.8 \\
    Ministral-8B & \textbf{25.8} & 21.0 & 6.7 & 58.8 & \textbf{69.3} & 44.7 & 23.8 & \textbf{25.8} & 15.6 & 35.4 & \textbf{39.2} & 28.4 \\
    Mistral-Nemo & \textbf{24.2} & 20.1 & 5.7 & \textbf{65.0} & 58.5 & 44.7 & \textbf{22.6} & 21.8 & 15.2 & 31.4 & \textbf{31.6} & 26.0 \\
    Mistral-Small & \textbf{31.8} & 23.1 & 6.6 & \textbf{72.6} & 70.1 & 45.5 & \textbf{31.2} & 26.2 & 16.8 & \textbf{41.2} & 38.6 & 29.8 \\
    gemma-2-2b & 9.9 & \textbf{11.0} & 6.4 & 42.5 & 34.5 & \textbf{46.8} & \textbf{12.6} & 11.0 & 7.2 & \textbf{17.8} & 14.8 & 16.2 \\
    gemma-2-9b & \textbf{39.8} & 37.5 & 14.3 & 67.9 & 61.1 & \textbf{70.8} & \textbf{23.2} & 22.4 & 17.2 & \textbf{35.6} & 33.0 & 34.8 \\
    gemma-2-27b & 27.6 & \textbf{54.4} & 23.7 & \textbf{83.2} & 67.5 & 76.9 & 25.8 & \textbf{27.2} & 23.6 & 40.4 & \textbf{41.2} & 37.0 \\
    Phi-3-mini & \textbf{22.5} & \textbf{22.5} & 20.1 & 51.6 & \textbf{71.9} & 59.1 & 21.0 & \textbf{21.6} & 22.0 & 30.8 & \textbf{36.8} & 31.2 \\
    Phi-3.5-mini & \textbf{15.2} & 13.0 & 6.7 & \textbf{58.8} & 54.0 & 40.2 & \textbf{20.0} & 14.8 & 15.2 & \textbf{31.4} & 29.2 & 30.2 \\
    Phi-3-small & \textbf{47.4} & 41.2 & 22.5 & \textbf{78.5} & 76.6 & 71.6 & \textbf{30.4} & 30.2 & 27.0 & \textbf{40.4} & 39.0 & 39.4 \\
    Phi-3-medium & \textbf{31.5} & 27.1 & 10.3 & \textbf{70.0} & 58.6 & 39.5 & \textbf{28.0} & 26.4 & 25.2 & 36.8 & 35.8 & \textbf{37.4} \\
    Phi-4 & \textbf{47.1} & 40.2 & 30.3 & \textbf{88.0} & 68.6 & 51.2 & \textbf{40.0} & 33.8 & 31.0 & \textbf{52.2} & 41.6 & 39.0 \\
    Llama-3.2-1B & \textbf{3.4} & 3.2 & 3.3 & 28.0 & \textbf{28.7} & 28.1 & 8.8 & 8.4 & \textbf{9.4} & \textbf{17.4} & 14.6 & 15.8 \\
    Llama-3.2-3B & \textbf{19.6} & 15.8 & 10.8 & 54.1 & 38.9 & \textbf{55.6} & \textbf{18.2} & 15.4 & 16.8 & \textbf{29.6} & 23.4 & 25.8 \\
    Llama-3.1-8B & \textbf{35.2} & 21.8 & 12.1 & \textbf{69.8} & 51.3 & 58.9 & \textbf{22.4} & 17.4 & 16.6 & \textbf{30.8} & 25.2 & 26.0 \\
    Llama-3.1-70B & \textbf{51.7} & 42.1 & 23.2 & \textbf{82.5} & 69.1 & 67.9 & \textbf{34.0} & 27.0 & 25.0 & \textbf{45.2} & 45.0 & 41.0 \\
    \bottomrule
    \end{tabular}
    }
    \caption{In most cases, c2f and aav prompt styles outperform sbs under strict token budgets. The highest accuracy for each model at each budget is highlighted in bold.}
    \label{tab:findings2}
\end{table*}

\subsection{Early Stopping Method Outperforms Direct Terminating}
\label{sec:findings 1}

\begin{tcolorbox}[colframe=blue!40, colback=white, coltitle=black,fonttitle=\bfseries, notitle]
    \textbf{Finding 1:} Compared to directly terminating at token budget, early stopping consistently improves LLMs’ reasoning performance on all combinations of datasets (GSM8K and MATH500) and prompt styles (sbs, c2f, and aav) in our evaluation.
\end{tcolorbox}

We illustrate the performance of both methods from three models in Figure \ref{fig:findings1_plot}, and plot results of all tested models in Appendix \ref{sec:appendix findings 1}. From these figures, we find the advantage on performance of early stopping method can last till its accuracy converges or crosses with the accuracy of directly terminating method. We also notice the difference of two methods on models' convergent accuracy. On GSM8K dataset, it is negligible, however on MATH500 this difference can be up to 5\%.

Terminating exactly at token budget truncates LLMs' reasoning process, leading to none or incorrect answer extraction. However, early stopping and concluding method can help LLMs to generate correct final answers even when partial reasoning steps are available
Therefore, in order to fully illustrate and study LLMs' reasoning capabilities under token budgets, we only report the results of early stopping method in the rest of this paper.

\subsection{Prompt and Thinking Pattern Matters}
\label{sec:findings 2}
\begin{tcolorbox}[colframe=blue!40, colback=white, coltitle=black,fonttitle=\bfseries, notitle]
    \textbf{Finding 2:} While the one-fits-all optimal prompt style doesn't exist, coarse-to-fine (c2f) and answer-and-verify (aav) outperform step-by-step (sbs) in most scenarios under output length constraint.
\end{tcolorbox}

Table \ref{tab:findings2} demonstrates the superiority of aav and c2f styles at different token budgets. Figures of complete results can be found in Appendix \ref{sec:appendix findings 2}. 
Combining with the examination of LLMs' responses, we surmise that the better LLMs can understand and follow the implication in prompts and the more lengthy reasoning steps they tend to generate, the more likely c2f and aav styles can help increase LLMs' performance than sbs style.

For example, we find that Qwen2.5-Math models can not follow c2f and aav formatted style, thus generating reasoning steps very much like sbs style.
DeepSeek-R1-Distill models generate responses in the correct style, but they all start with a thinking trace wrapped between <think> and </think>. Therefore, we observe negligible improvement on these models when switching prompts to encourage models to output brief analysis or speculative answers at early stage of their reasoning process.

Compared with other models on the same dataset, QwQ, Qwen-2.5 (7, 14, 32B) and Phi-4 models have more performance improvement when prompted in c2f or aav style. We assume this stems from the fact that they tend to generate more reasoning steps, thus more token budget is needed to achieve accuracy convergence when prompted in sbs style (over 400 tokens on GSM8K and 1k tokens on MATH500). Therefore, their responses may have more redundancy and can be compressed in a more concise way when prompted in c2f or aav style, which results in higher possibility of correct answer derivation under token budget.

More difficult dataset, MATH500, requires more lengthy and more \textit{informative} steps to get the problems solved. So the benefits of answering questions prematurely by compressing reasoning steps will decrease, which leads to limited improvement of c2f and aav styles over sbs.

\begin{figure}[htbp]
    \includegraphics[width=\columnwidth]{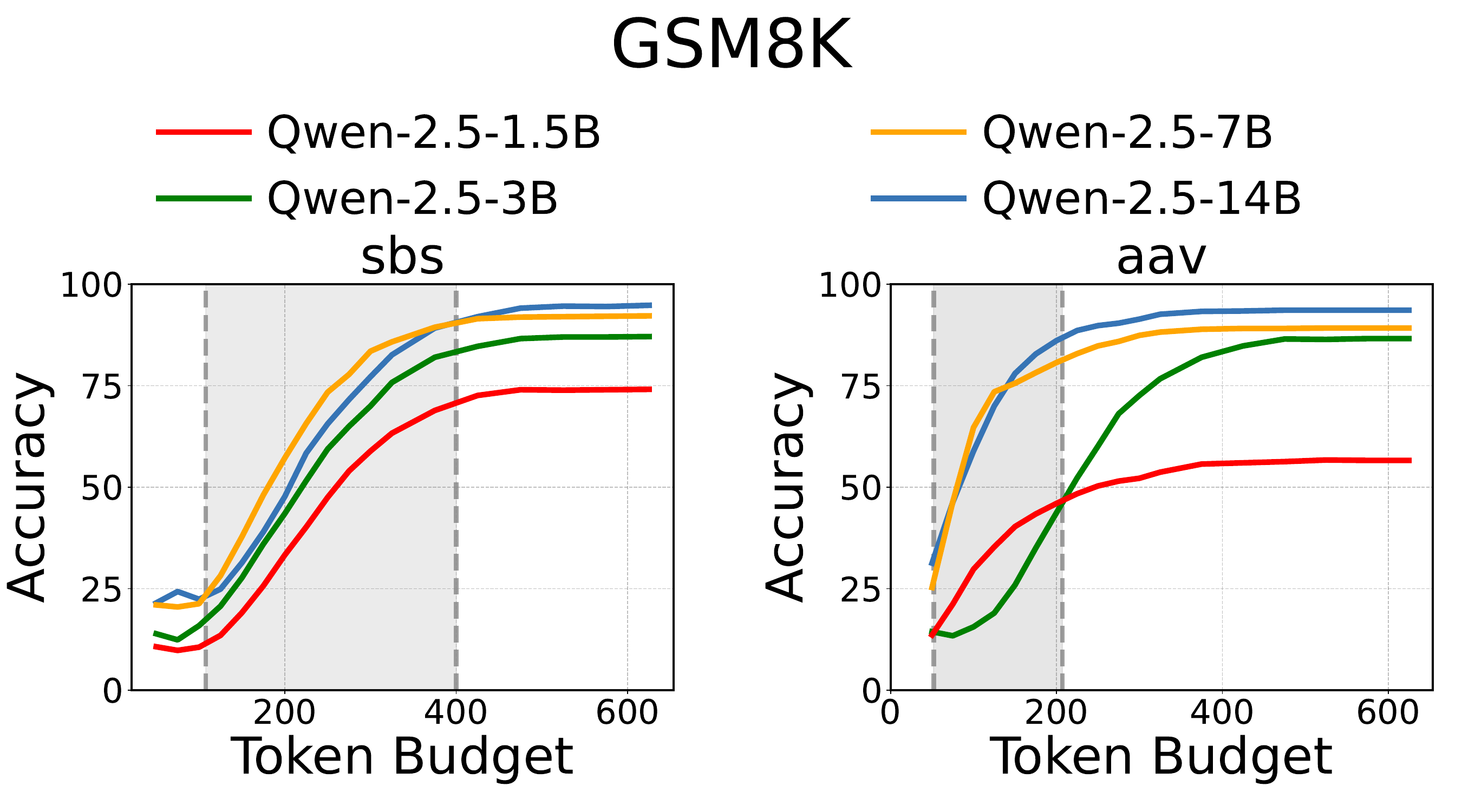}
    \includegraphics[width=\columnwidth]{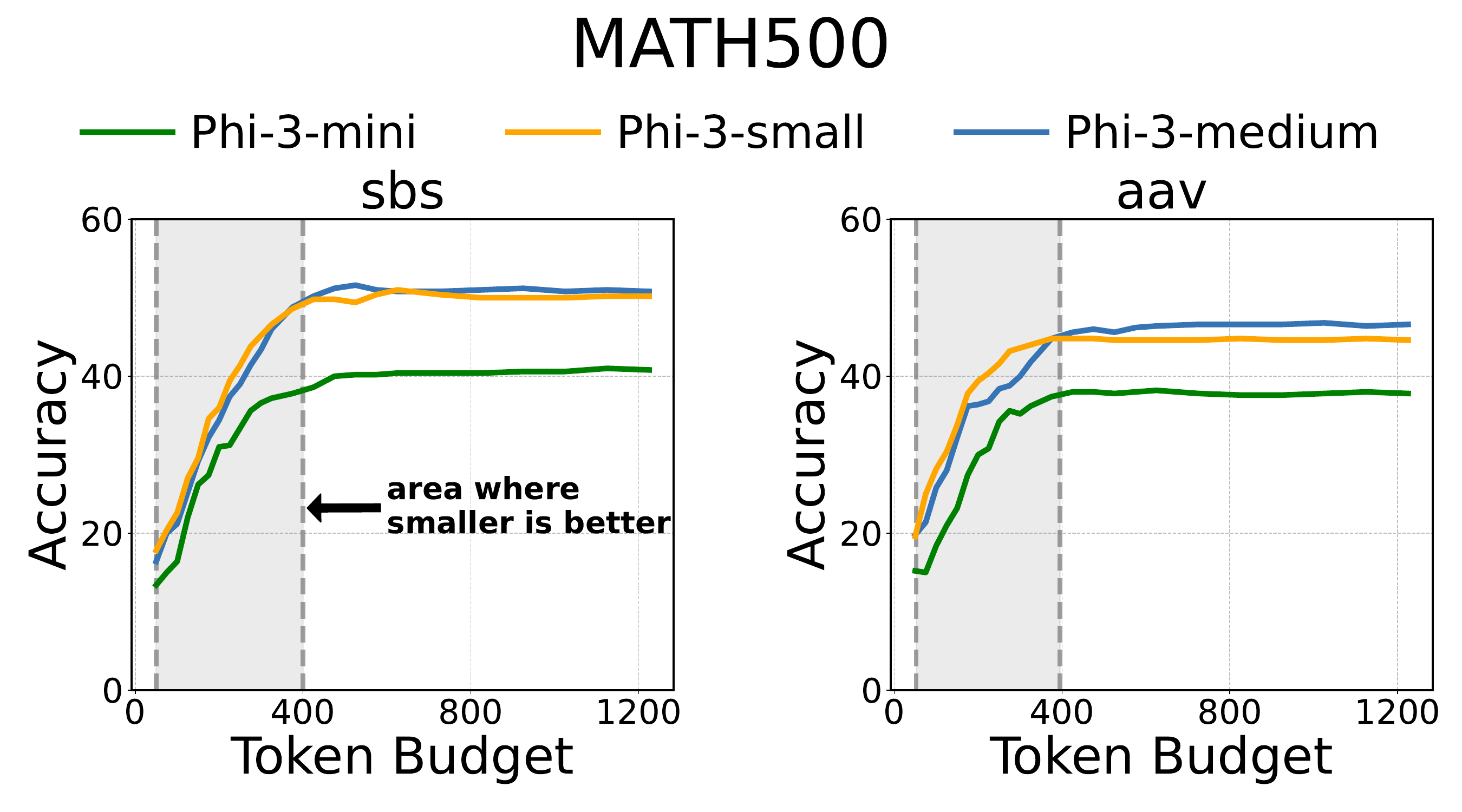}
    \vspace{-0.4cm}
    \caption{Under token budget, small models can outperform larger models on both datasets.}
    \label{fig:findings3_plot}
\end{figure}

\subsection{Larger Models are NOT Always Better}
\label{sec:findings 3}

\begin{tcolorbox}[colframe=blue!40, colback=white, coltitle=black,fonttitle=\bfseries, notitle]
    \textbf{Finding 3:} Under strict output constraints, the reasoning performance of LLMs might not scale monotonically with the model size. In other words, larger is not always better.
\end{tcolorbox}

\begin{figure*}[!htbp]
    \centering
    \includegraphics[width=\textwidth, trim=50 25 0 10, clip]{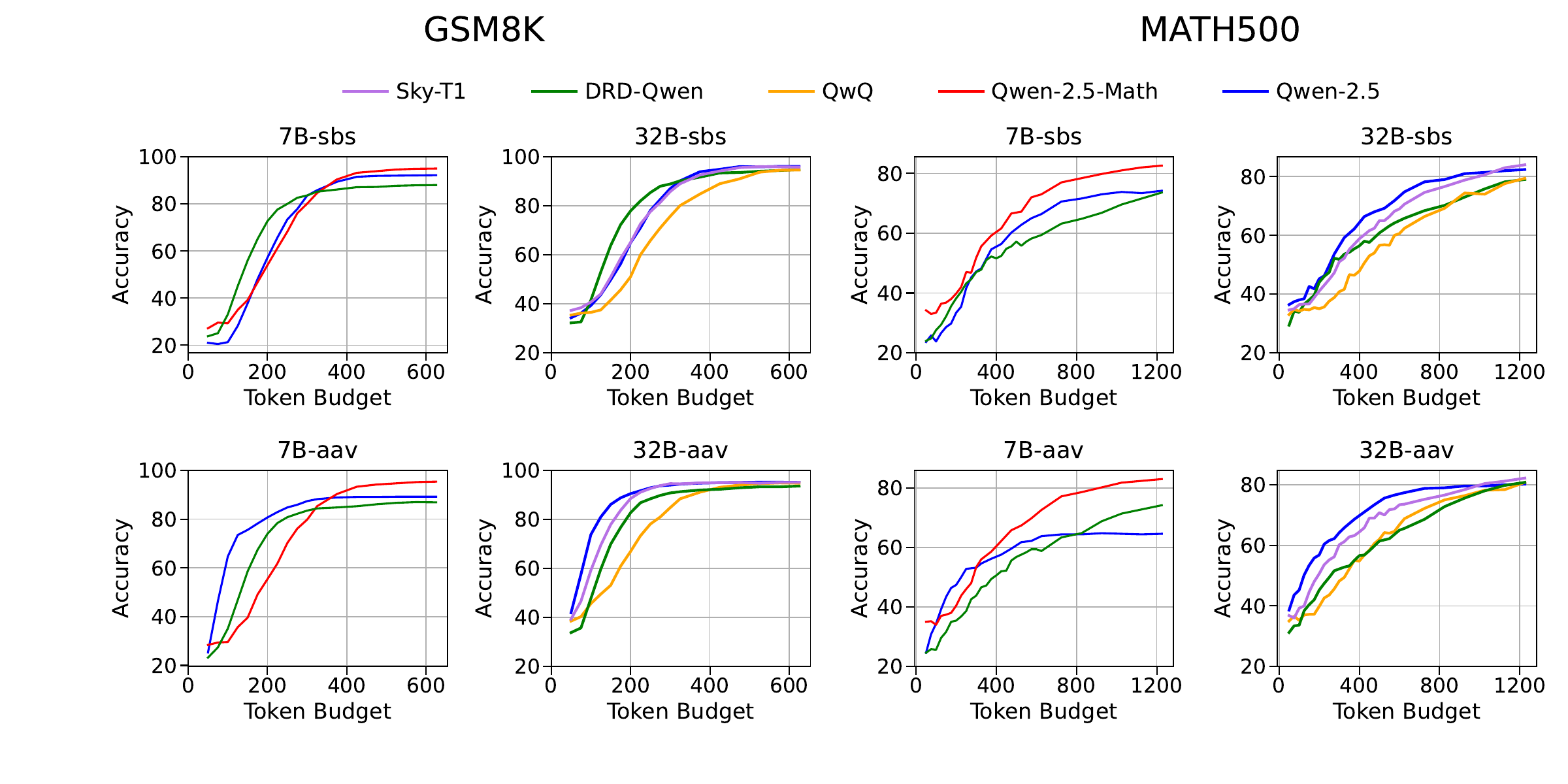}
    \vspace{-0.8cm}
    \caption{Under token budgets, reasoning models are not always better than instruction tuned or math models.} 
    \label{fig:findings4_plot}
\end{figure*}

The most representative examples for this finding are the Qwen-2.5 and Phi-3 series.
As shown in Figure \ref{fig:findings3_plot}, the highlighted parts in gray background is the area where anomalies exist. On the GSM8K dataset, within the token budget from 100 to 400 and prompted in sbs style, the accuracy of Qwen-2.5-7B model is consistently higher than 14B. Similar phenomenon also exists in aav prompt style under token budget of 200 for Qwen-2.5-7B versus 14B, and Qwen-2,5-1.5B versus 3B.
As for Phi-3 series on MATH500, 7B (Phi-3-small) model matches or even surpass 14B (Phi-3-medium) model under token budget of 400. Considering the 50\% saving in size of smaller models from above pairwise comparison, their advantage over larger models is quite surprising. We plot all examples in Appendix \ref{sec:appendix findings 3}.

After checking the responses of Qwen models, we find that smaller Qwen models (7B/1.5B) require fewer output tokens to complete GSM8K problems than their larger counterparts (14B/3B). For instance, the median and average output token length for 7B model are 14.6\% and 14.2\% smaller than 14B, tested using sbs style and directly terminating method with budget set to 4096. 
On GSM8K, both small and large models can achieve high accuracy of completed questions, but they are prone to make mistakes when forced to early stop and conclude. Thus large models perform worse than small ones under strict token budgets.

One possible explanation for the abnormal phenomenon of Phi-3 model series can be found in its technical report \citep{abdin2024phi3technicalreporthighly}, which states that Phi-3-medium model is trained on the same amount of data with Phi-3-small but for slightly more epochs. The improvement from 7B to 14B is not as significant as that from 3.8B to 7B on several benchmarks. Our experiments from Table \ref{tab:all_models}, Figure \ref{fig:findings1_gsm8k_all_aav_plot} and Figure \ref{fig:findings1_math500_all_aav_plot} also indicate that they have similar capability on both GSM8K and MATH500 datasets. This implies that trained on the same amount of data, LLM's reasoning ability under strict token budget may be diluted by over-large model size. 

Finding 3 inspires us to reconsider the relationship between LLMs' reasoning capability and parameter size. Although larger models exhibit stronger capabilities on most benchmarks under no output limit, they may underperform smaller models on the ability to conduct precise and concise reasoning under strict constraint of token budgets.

\subsection{Reasoning Models are NOT Always Better}
\label{sec:findings 4}

\begin{tcolorbox}[colframe=blue!40, colback=white, coltitle=black,fonttitle=\bfseries, notitle]
    \textbf{Finding 4:} Under strict output length constraints, reasoning models don't always outperform instruction tuned or math models.
\end{tcolorbox}

In Figure \ref{fig:findings4_plot}, on GSM8K, DRD-Qwen models (7B, 32B) perform the best within token budgets smaller than 300, prompted in sbs style. But when using aav style, Qwen-2.5 instruction tuned models become the best choices, as they output speculated answers at early stage of reasoning, which helps to conclude final answers under strict token budgets. 
On MATH500, Qwen2.5-Math-7B model performs the best among models of 7B size under token budgets between 300 and 1.2k. And Qwen-2.5 32B instruction tuned model consistently performs the best among models of 32B size within token budget of 1000, prompted in both sbs and aav styles.

For easy questions, LLMs can solve them using a few simple reasoning steps, thus the conciseness and length of responses are more important. Therefore, instruction tuned models can beat reasoning models when prompted with aav style to derive correct answers more efficiently. For complex questions, reasoning models tend to generate much longer reasoning steps than other models, which can be identified from Figure \ref{fig:findings1_plot} and Appendix \ref{sec:appendix findings 1}. Although reasoning models can achieve higher accuracies when token budget is large enough, their accuracy curves rise more slowly at early stages and can not match those of instruction tuned or math models under strict output token budgets.

\begin{figure}[htbp]
    \centering
    \includegraphics[width=0.49\columnwidth]{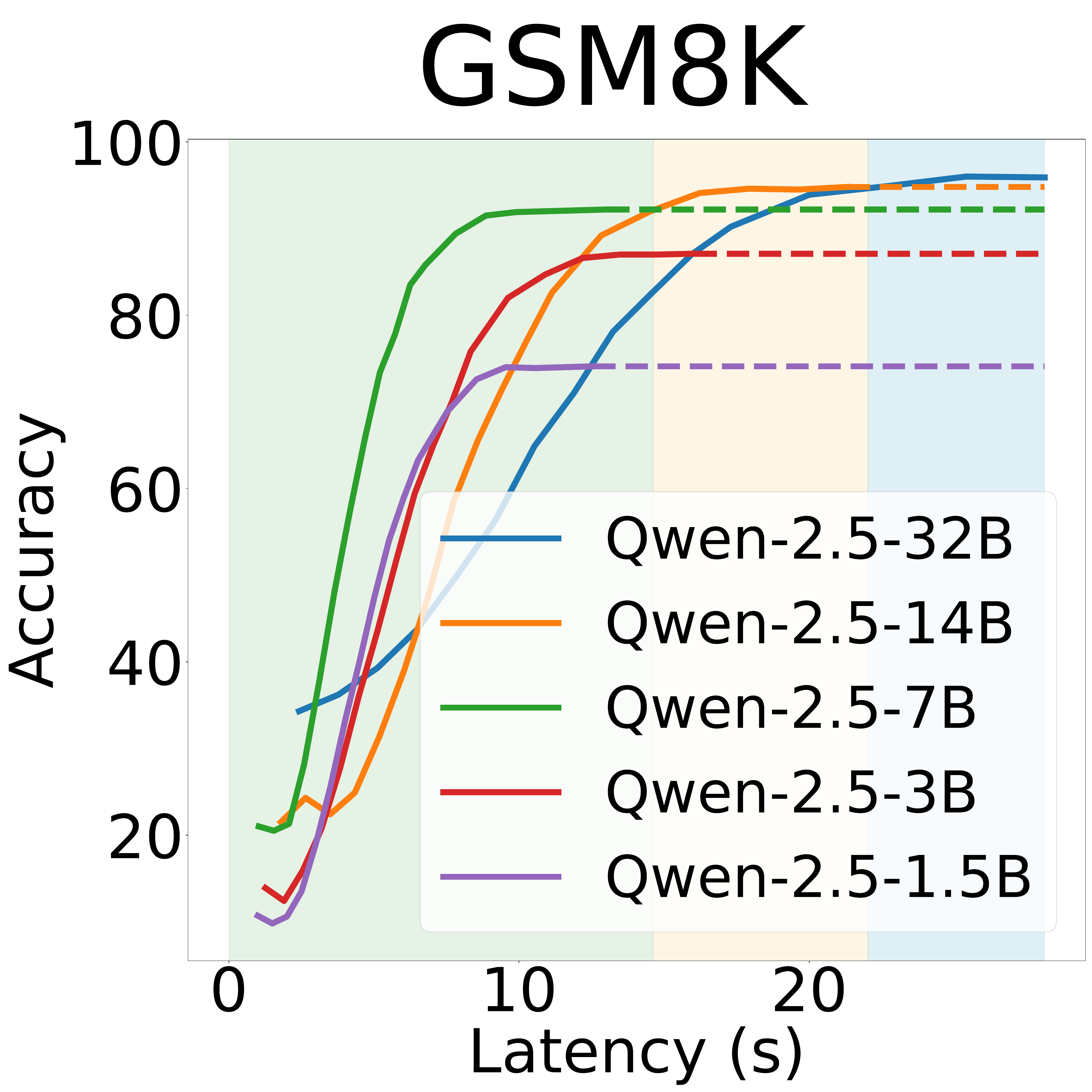} 
    \hfill
    \includegraphics[width=0.49\columnwidth]{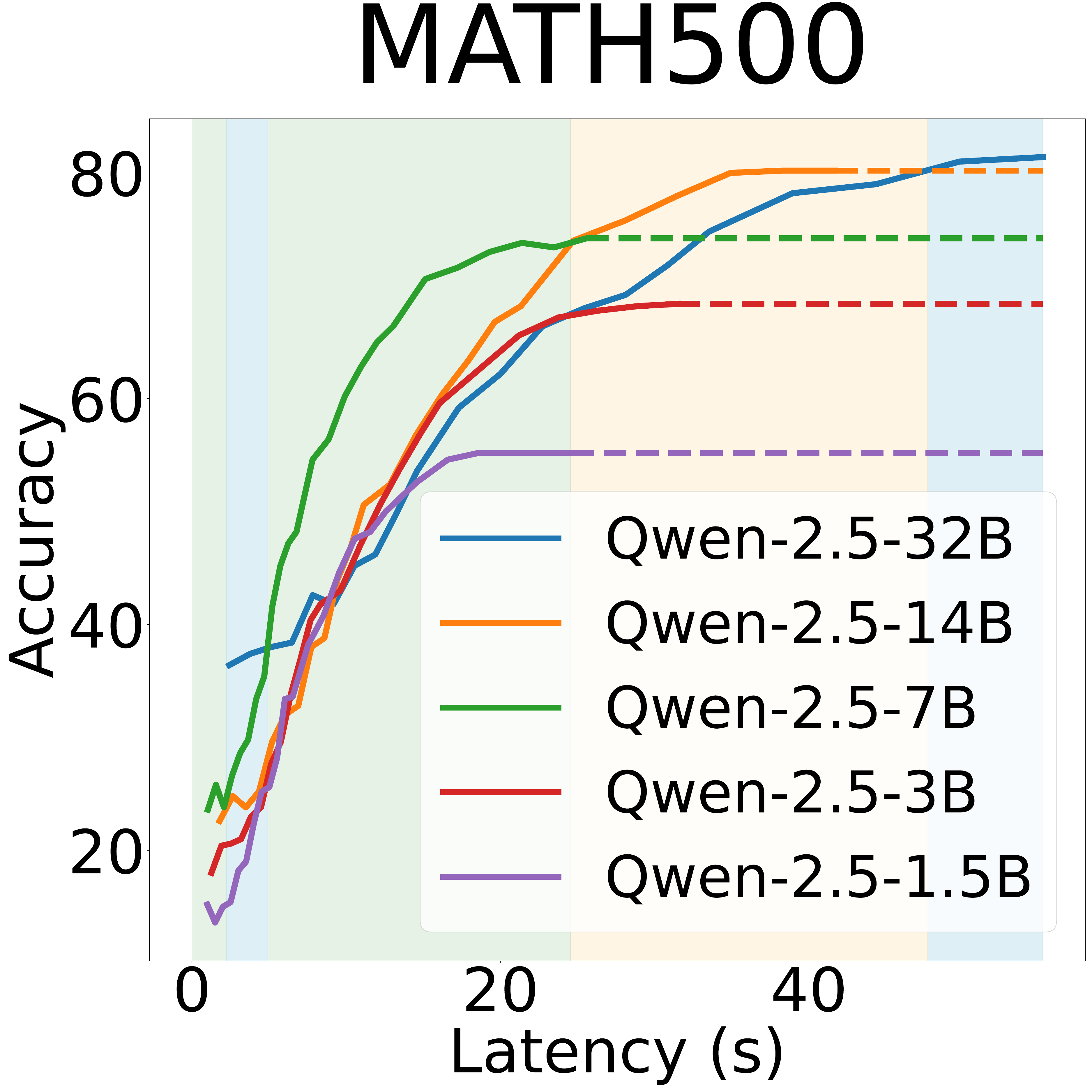}
    \vspace{-0.7cm}
    \caption {Comparison of Qwen-2.5-Instruct models on NVIDIA A800 GPU under infernce latency budget.}
    \label{fig:findings5_plot}
\end{figure}

\subsection{Mid-Sized Models are Latency-Optimal}
\label{sec:findings 5}

\begin{tcolorbox}[colframe=blue!40, colback=white, coltitle=black,fonttitle=\bfseries, notitle]
    \textbf{Finding 5:} In latency critical scenarios that require on-device reasoning, we should prioritize
    middle-sized models, even if there is enough resource to deploy a larger model.
\end{tcolorbox}

Given the specific device to deploy, the inference latency of LLMs is mainly composed of prefill time and decode time. The prefill latency is determined by model input length, and decode latency is the sum of latency of every new token, which is determined by the context length before generating each token. We first evaluate the input length for all scenarios in Table \ref{tab:input token count} from Appendix \ref{sec:appendix findings 5}, which shows that the input ranges within 150-250 tokens. Then we measure the correlation between latency and input and output length for all models and prompt styles on a single NVIDIA A800 GPU. From examples in Figure \ref{fig:token_latency} we find that the impact of input length on total latency can be ignored. And for each model, there exists a clear linear mapping between output tokens and inference latency.

To find the optimal model size under latency budgets, we plot the performance of Qwen-2.5-Instruct models prompted in sbs style in Figure \ref{fig:findings5_plot}, using the latency mapping derived through on-device profiling.
The background of each region is highlighted using the curve color of the model that dominates that area on accuracy.
On both datasets, 7B model exhibits significant advantage within limited latency budgets. As the budget relaxes, the performance of 14B and 32B models surpasses 7B model.
Results of other model series can be found in Figure \ref{fig:findings5_gsm8k_all_plot} and Figure \ref{fig:findings5_math_all_plot}, where the general trend is similar. This finding implies when deploying LLMs on devices under strict latency constraints, we should prioritize middle sized models, with a typical value around 7B.

\section{Discussion }

Apart from math reasoning tasks, we also cautiously speculate that our findings can generalize to other similar reasoning domains.
So we conduct extra experiments of Qwen-2.5 and DRD-Qwen model series on mmlu\_stem and ACPBench \citep{kokel2024acp} datasets. mmlu\_stem is a subset of STEM subjects (such as astronomy and biology) defined in MMLU \citep{hendryckstest2021}. ACPBench contains both single and multi step reasoning tasks for evaluating actions and plans.

From experiment results, we have observed the performance improvement when switching sbs prompt style into aav or c2f for Qwen-2.5 models on mmlu\_stem dataset, which adheres to our finding 2. However, this improvement is rather limited for Qwen-2.5 models on ACPBench. As for different model sizes, we find that Qwen-2.5 14B can outperform 32B on ACPBench when prompted with aav style. Qwen-2.5 1.5B also outperforms 3B on ACPBench when prompted with c2f. These results support our finding 3 that large models are not always better than smaller ones. 
We also compare the impact of model types of the same size.
On both datasets, we notice notable advantage of instruction tuned models under token budgets within 1000, although reasoning models have higher accuracy when budget is relaxed to 4096.

\section{Conclusion}

We investigated the reasoning capabilities of large language models (LLMs) under time-constrained scenarios by imposing output token length limitations. 
Our findings reveal that the performance of LLMs can vary significantly depending on factors such as model size, architecture, and prompt design when operating under different constraints. We expect this work to shed light on this under-explored area and offer valuable insights for practitioners aiming to deploy LLMs in real-world and time-sensitive applications.

\section*{Limitations}

While our study provides a first step toward understanding LLM reasoning under time-constrained conditions, it has several limitations. First, we focused primarily on mathematical reasoning tasks, which, while representative, may not fully capture the diverse range of real-world applications where time constraints are critical. Future work should extend this analysis to other domains, such as programming and decision making. Second, our experiments were conducted using a limited set of LLMs and prompt designs, which may not comprehensively represent the broader landscape of available models and techniques. Third, the validity of our findings (e.g. the optimal model sizes under different time budgets) may be threatened by the different training procedures of each model, which are usually not transparent to researchers. Finally, our evaluation assumes a direct correlation between token budget and latency, which, though practical, does not account for hardware-specific variations in real-world deployment. 


\bibliography{references}

\appendix
\section{Method Details}
\subsection{Answer Extraction and Scoring}
\label{sec:appendix answer extraction and scoring}

Answer extraction relies on several pattern matching operations. This process searches for specific formats, and in most cases LLMs will put their final answers within \textbackslash\textbackslash boxed\{\}. Other phrases like "the answer is", "final answer is" are also used to locate the answer after them. If no specific patterns are found, the default method is to extract the last numeric value from the string, which guarantees that answers can be extracted even when the format does not strictly adhere to predefined patterns.

After the initial extraction, the function undergoes several post-processing steps to clean and normalize the extracted answer. These steps include removing leading colons, trailing periods, and slashes. Moreover, the extracted answer is cleaned of unnecessary whitespace and, depending on the dataset, may have units removed.

Finally, results from some models are manually reviewed to correct any formatting errors or inconsistencies that may have been overlooked during the automated extraction and post-processing steps. This manual review process enhances the accuracy and reliability of the extracted answers, particularly in cases where the output string does not strictly follow expected patterns or where the automated extraction process might have introduced errors.

After the answer is extracted and cleaned, we use scoring method from framework Qwen-2.5-Math \citep{yang2024qwen25mathtechnicalreportmathematical}. The accuracy is calculated based on both numerical and symbolic equality between the extracted answer and labeled answer from datasets.

\subsection{Token Budget Implication and Concluding Message}
\label{sec:appendix implication and concluding}

\begin{figure}[!htbp]
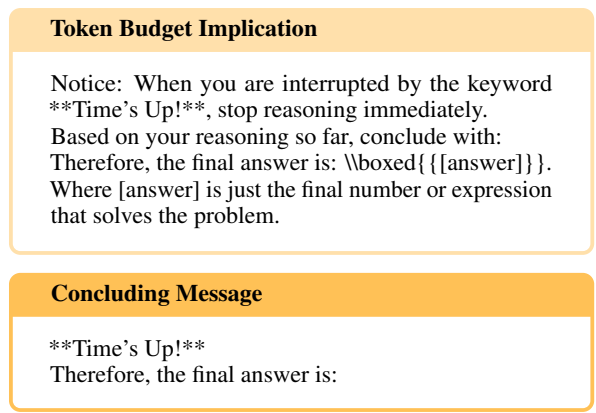

    \centering
    \begin{tcolorbox}[colframe=orange!40, colback=white, coltitle=black,fonttitle=\small\bfseries, title={Token Budget Implication}, fontupper=\small]
        Notice: When you are interrupted by the keyword **Time’s Up!**, stop reasoning immediately. \\
        Based on your reasoning so far, conclude with: \\
        Therefore, the final answer is: \textbackslash\textbackslash boxed\{\{[answer]\}\}.\\
        Where [answer] is just the final number or expression that solves the problem.
    \end{tcolorbox}

    \begin{tcolorbox}[colframe=orange!80, colback=white, coltitle=black,fonttitle=\small\bfseries, title={Concluding Message}, fontupper=\small]
       **Time's Up!**\\
       Therefore, the final answer is:
    \end{tcolorbox}
    \caption{Token budget implication and concluding message used in early stopping method.}\
    \label{fig:implication}
\end{figure}

Here is the full version of token budget implication and concluding message used in early stopping method. Phrase \verb|**Time's Up!**| is used as the signal to execute early stop, as shown in Figure \ref{fig:implication}.

\section{Experiment Setup Details}
\label{sec:appendix experiment setup}

\subsection{Prompt Style}
\label{sec:appendix prompt style}

We list the full version of step-by-step (sbs), coarse-to-fine (c2f), and answer-and-verify (aav) prompt styles in Figure \ref{fig:prompt styles}. Step-by-step style requires the LLM to follow a direct linear reasoning process. 
Coarse-to-fine starts with a brief initial answer (coarse) and then adds detailed reasoning (fine). 
Answer-and-verify begins with an intuitive answer, followed by verification and correction through detailed reasoning.

\begin{figure}
\begin{tcolorbox}[colframe=gray!40, colback=white, coltitle=black,fonttitle=\small\bfseries, title={Step by step (sbs) prompt style}, fontupper=\small]
    Please reason step by step. \\
    Conclude with: \\
    Therefore, the final answer is: \textbackslash\textbackslash boxed\{\{[answer]\}\}.\\
    Where [answer] is just the final number or expression that solves the problem.
\end{tcolorbox}

\begin{tcolorbox}[colframe=gray!40, colback=white, coltitle=black,fonttitle=\small\bfseries, title={Coarse to fine (c2f) prompt style}, fontupper=\small]
   Use the following pattern to solve the problem:\\
**Coarse-Grained Reasoning**\\
Provide a brief analysis and initial answer, focusing on efficiency and conciseness.\\ \\
**Fine-Grained Reasoning**\\
Provide detailed reasoning step by step and a refined answer, focusing on correctness and rigor.\\

Conclude with: \\
Therefore, the final answer is: \textbackslash\textbackslash boxed\{\{[answer]\}\}.\\
Where [answer] is just the final number or expression that solves the problem.
\end{tcolorbox}

\begin{tcolorbox}[colframe=gray!40, colback=white, coltitle=black,fonttitle=\small\bfseries, title={Answer and verify (aav) prompt style}, fontupper=\small]
  Use the following pattern to solve the problem:\\
**Quick Answer**\\
Provide an initial answer based on intuition or quick calculation.\\\\
**Verification**\\
Provide a revised answer through reasoning step by step. Correct previous mistakes, if any.\\\\
Conclude with: \\
Therefore, the final answer is: \textbackslash\textbackslash boxed\{\{[answer]\}\}.\\
Where [answer] is just the final number or expression that solves the problem.
\end{tcolorbox}

\caption{Three different prompt styles used in experiments.}
\label{fig:prompt styles}
\end{figure}

\begin{figure*}[!htbp]
    \centering
    \begin{tcolorbox}[colframe=orange!40, colback=white, coltitle=black,fonttitle=\small\bfseries, title={Chat Templates}, fontupper=\small]
    \setlength{\baselineskip}{1.5em}
    "mistral\_format": "<s>[INST] \{system\_message\}\textbackslash n\textbackslash n\{input\}[/INST]",
    
    "qwen\_format": "<|im\_start|>system\textbackslash n\{system\_message\}<|im\_end|>\textbackslash n<|im\_start|>user\textbackslash n\{input\}<|im\_end|>\textbackslash n<|im\_start|>\\
    assistant\textbackslash n",
    
    "phi3mini\_format": "<|system|>\textbackslash n\{system\_message\}<|end|>\textbackslash n<|user|>\textbackslash n\{input\}<|end|>\textbackslash n<|assistant|>\textbackslash n",
    
    "phi3small\_format": "<|endoftext|><|system|>\textbackslash n\{system\_message\}<|end|>\textbackslash n<|user|>\textbackslash n\{input\}<|end|>\textbackslash n<|assistant|>\textbackslash n",
    
    "phi3medium\_format": "<|user|>\textbackslash n\{input\}<|end|>\textbackslash n<|assistant|>\textbackslash n",
    
    "phi4\_format": "<|im\_start|>system<|im\_sep|>\{system\_message\}<|im\_end|><|im\_start|>user<|im\_sep|>\{input\}<|im\_end|>\\
    <|im\_start|>assistant<|im\_sep|>",
    
    "llama\_format": "<|begin\_of\_text|><|start\_header\_id|>system<|end\_header\_id|>\textbackslash n\textbackslash n\{system\_message\}<|eot\_id|>\\
    <|start\_header\_id|>user<|end\_header\_id|>\textbackslash n\textbackslash n\{input\}<|eot\_id|><|start\_header\_id|>assistant<|end\_header\_id|>\textbackslash n\textbackslash n",
    
    "gemma\_format": "<bos><start\_of\_turn>user\textbackslash n\{input\}<end\_of\_turn>\textbackslash n<start\_of\_turn>model\textbackslash n",
    
    "deepseek-r1-distill\_format" : "<|begin\_of\_sentence|><|User|>\{input\}<|Assistant|>"
    \end{tcolorbox}
    \caption{Chat templates of different model series tested in our experiments.}
    \label{fig:chat_templates}
\end{figure*}

\subsection{Model Input Construction}
\label{sec:appendix model input}
\begin{figure*}[htbp]
    \centering
    \includegraphics[width=\textwidth]{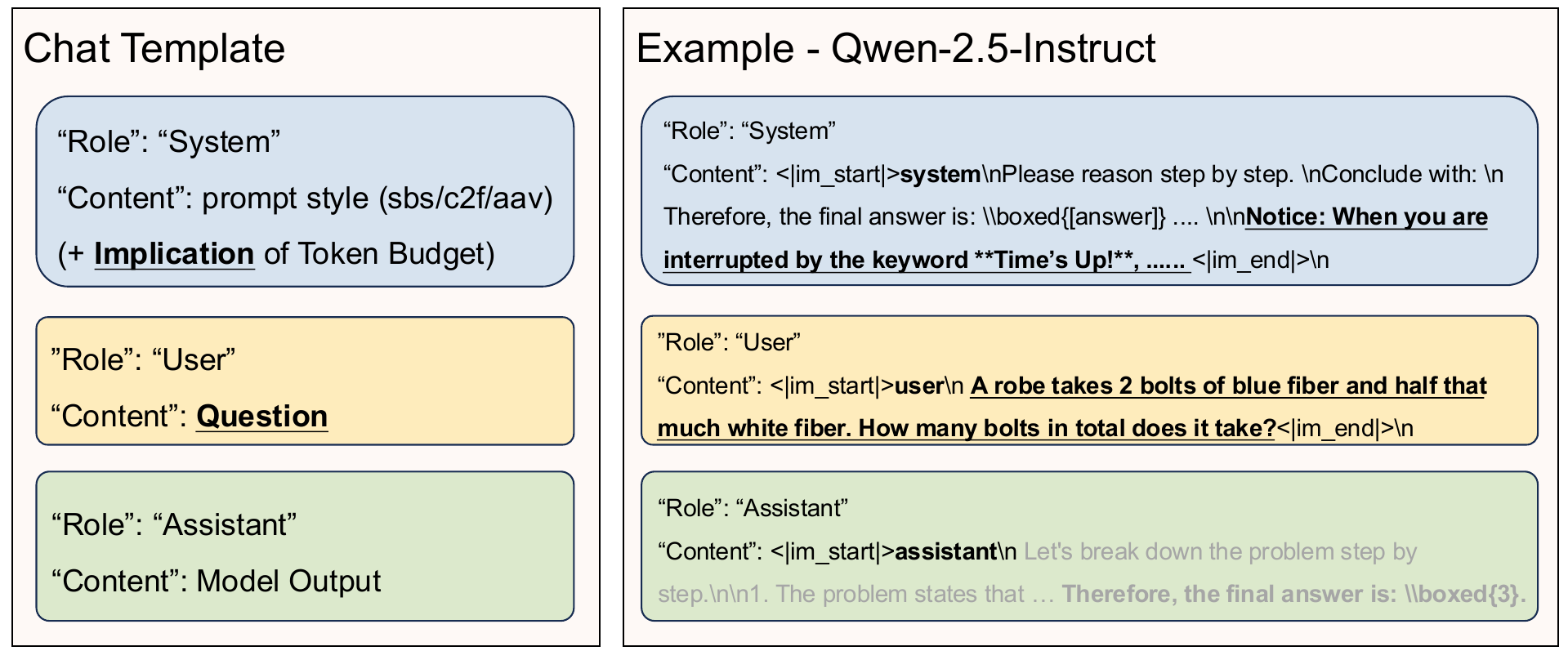}
    \caption {The chat template and an example of constructing model input. }
    \label{fig:model_input}
\end{figure*}

Figure \ref{fig:chat_templates} shows the chat templates we use for all models in our experiments. The \{system\_message\} will be replaced by prompt styles like sbs, c2f or aav, and \{input\} will be replaced by math problems. If the template does support system role, we concatenate prompt styles and problem with "\textbackslash n\textbackslash n" and then feed them into \{input\}. 
Figure \ref{fig:model_input} shows an example of constructing model input based on Qwen-2.5 chat templates and sbs prompt style.

\subsection{Discussion About License or Terms for Scientific Artifacts}
\label{sec: appendix license}
All datasets, models and evaluation framework in this work are strictly used for academic and research purposes only and do not involve any commercial applications. Their usage is fully compliant with their respective licenses and the intended use specified by the original providers. No modifications or derivatives of them have been used in ways that would conflict with the terms set forth by the original licenses. The data and models have been used solely within the scope of this research and will not be deployed in any commercial or non-research contexts.

\section{Evaluation Details}
\subsection{Finding 1}
\label{sec:appendix findings 1}

Here we present a comparative analysis of various methods utilizing direct terminating (dashed lines) and early stopping (solid lines) on GSM8K and MATH500 datasets. The methods are evaluated across different prompting styles: step-by-step (sbs, Figure \ref{fig:findings1_gsm8k_all_sbs_plot},\ref{fig:findings1_math500_all_sbs_plot}), coarse-to-fine (c2f, Figure \ref{fig:findings1_gsm8k_all_c2f_plot},\ref{fig:findings1_math500_all_c2f_plot}), and answer-and-verify (aav, Figure \ref{fig:findings1_gsm8k_all_aav_plot},\ref{fig:findings1_math500_all_aav_plot}). Each prompting style illustrates the performance differs from different models under varying prompting strategies, showing which strategies are more effective on specific datasets.

\begin{figure*}[htbp]
    \centering
    \includegraphics[width=\textwidth, trim=10 10 10 10, clip]{figures/findings1/gsm8k_all_sbs.pdf}
    \caption {Early-stopping (solid line) outperforms directly terminating (dashed line) method on GSM8K datasets. Prompting style: \textcolor{orange}{sbs}.}
    \label{fig:findings1_gsm8k_all_sbs_plot}
    \includegraphics[width=\textwidth, trim=10 10 10 10, clip]{figures/findings1/math500_all_sbs.pdf}
    \caption {Early-stopping (solid line) outperforms directly terminating (dashed line) method on MATH500 datasets. Prompting style: \textcolor{orange}{sbs}.}
    \label{fig:findings1_math500_all_sbs_plot}
\end{figure*}

\begin{figure*}[htbp]
    \centering
    \includegraphics[width=\textwidth, trim=10 10 10 10, clip]{figures/findings1/gsm8k_all_c2f.pdf}
    \caption {Early-stopping (solid line) outperforms directly terminating (dashed line) method on GSM8K datasets. Prompting style: \textcolor{green}{c2f}.}
    \label{fig:findings1_gsm8k_all_c2f_plot}
    \includegraphics[width=\textwidth, trim=10 10 10 10, clip]{figures/findings1/math500_all_c2f.pdf}
    \caption {Early-stopping (solid line) outperforms directly terminating (dashed line) method on MATH500 datasets. Prompting style: \textcolor{green}{c2f}.}
    \label{fig:findings1_math500_all_c2f_plot}
\end{figure*}

\begin{figure*}[htbp]
    \centering
    \includegraphics[width=\textwidth, trim=10 10 10 10, clip]{figures/findings1/gsm8k_all_aav.pdf}
    \caption {Early-stopping (solid line) outperforms directly terminating (dashed line) method on GSM8K datasets. Prompting style: \textcolor{purple}{aav}.}
    \label{fig:findings1_gsm8k_all_aav_plot}
    \includegraphics[width=\textwidth, trim=10 10 10 10, clip]{figures/findings1/math500_all_aav.pdf}
    \caption {Early-stopping (solid line) outperforms directly terminating (dashed line) method on MATH500 datasets. Prompting style: \textcolor{purple}{aav}.}
    \label{fig:findings1_math500_all_aav_plot}
\end{figure*}

\subsection{Finding 2}
\label{sec:appendix findings 2}

\begin{figure*}[htbp]
    \centering
    \includegraphics[width=\textwidth, trim=10 10 10 10, clip]{figures/findings2/gsm8k_all.pdf}
    \caption {Comparison of different models' performance with early-stopping methods on GSM8K datasets. Prompting style: \textcolor{orange}{sbs}, \textcolor{green}{c2f} and \textcolor{purple}{aav}.}
    \label{fig:findings2_gsm8k_all_plot}
    \includegraphics[width=\textwidth, trim=10 10 10 10, clip]{figures/findings2/math500_all.pdf}
    \caption {Comparison of different models' performance with early-stopping methods on MATH500 datasets. Prompting style: \textcolor{orange}{sbs}, \textcolor{green}{c2f} and \textcolor{purple}{aav}.}
    \label{fig:findings2_math500_all_plot}
\end{figure*}

Here we present a comparative analysis of various models' performance utilizing different prompting styles on the GSM8K (Figure \ref{fig:findings2_gsm8k_all_plot}) and MATH500 (Figure \ref{fig:findings2_math500_all_plot}) datasets. The models are evaluated across step-by-step solution (sbs), coarse-to-fine (c2f), and answer-and-verify (aav). Each prompting style illustrates how the performance differs between different models under these 3 prompting strategies.

\subsection{Finding 3}
\label{sec:appendix findings 3}

\begin{figure*}[htbp]
    \includegraphics[width=\columnwidth]{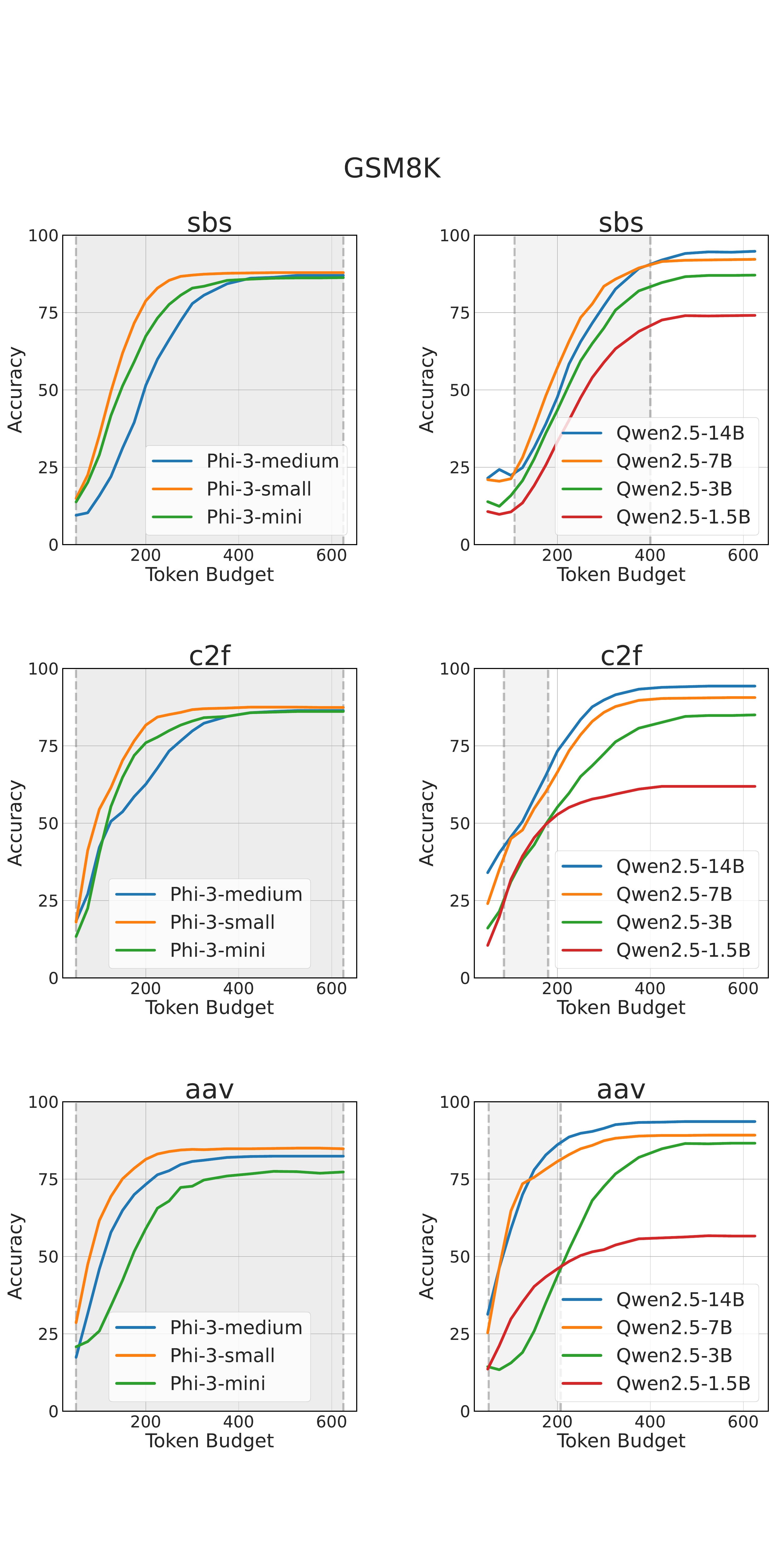}
    \hfill
    \includegraphics[width=\columnwidth]{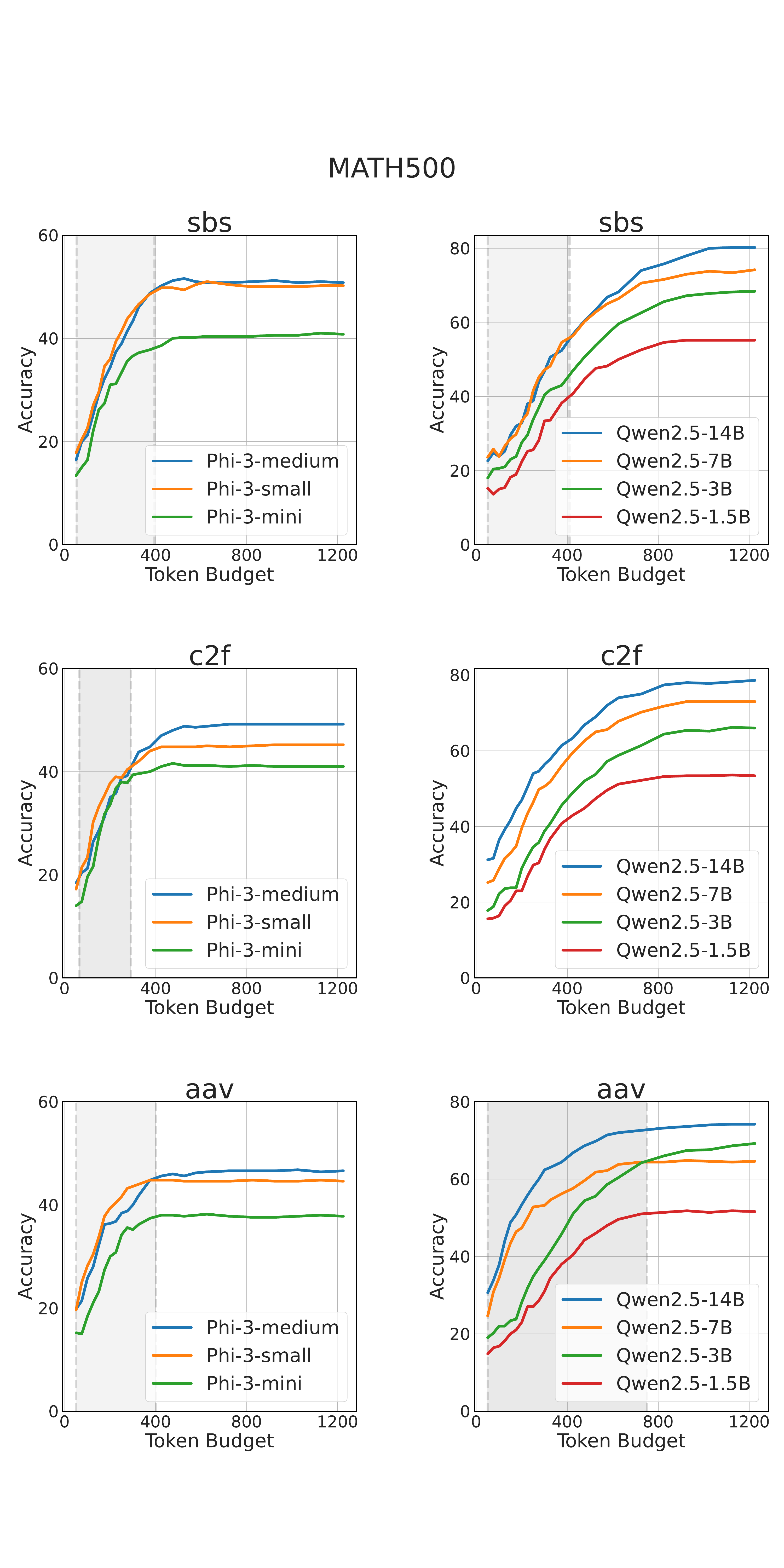}
    \caption{Qwen-2.5-Instruct and Phi-3 models' performance on GSM8K and MATH500 datasets with different token budgets.}
    \label{fig:hello}
\end{figure*}

Figure \ref{fig:hello} presents an analysis of the Qwen-2.5-Instruct and Phi-3 models' performance on the GSM8K and MATH500 datasets under different token budgets. This analysis explores how the reasoning performance of large language models (LLMs) may not scale monotonically with the model size under certain output constraints. In other words, larger models do not always perform better. The figures illustrate this phenomenon by showing the performance variations of the models with different token budgets on both datasets.

\subsection{Finding 5}
\label{sec:appendix findings 5}

\begin{table}[!htbp]
    \centering
    \small
    \setlength{\tabcolsep}{3pt}
    \resizebox{0.9\columnwidth}{!}{ 
    \begin{tabular}{c|cccccc}
    \toprule
    \multirow{2}{*}{Model Series} & \multicolumn{3}{c|}{GSM8K} & \multicolumn{3}{c}{MATH500} \\
    \cmidrule(lr){2 - 4} \cmidrule(lr){5 - 7}
     & sbs & c2f & aav & sbs & c2f & aav \\
    \midrule
    Mistral & 185 & 251 & 234 & 175 & 241 & 224 \\
    Qwen-2.5 & 170 & 222 & 207 & 163 & 215 & 200 \\
    Phi-3-mini & 184 & 254 & 235 & 172 & 242 & 223 \\
    Phi-3-small & 167 & 219 & 204 & 161 & 213 & 198 \\
    Phi-3-medium & 184 & 254 & 235 & 172 & 242 & 223 \\
    Phi-4 & 166 & 218 & 203 & 160 & 212 & 197 \\
    Llama-3.2 & 171 & 223 & 208 & 165 & 217 & 202 \\
    gemma-2 & 176 & 229 & 217 & 169 & 222 & 210 \\
    DRD-Qwen & 163 & 215 & 200 & 156 & 208 & 193 \\
    \bottomrule
    \end{tabular}
    }
    \caption{The median of input token counts for different model types across various datasets and prompt styles.}
    \label{tab:input token count}
\end{table}
Here we list the median of input token length for all models in various dataset and prompt style scenarios. Model inputs are formatted following the construction in Section \ref{sec:appendix model input}. For each model series, we use one of the models' tokenizer to encode all inputs.
The results are shown in Table \ref{tab:input token count}.

\begin{figure}[H]
    \includegraphics[width=\columnwidth]{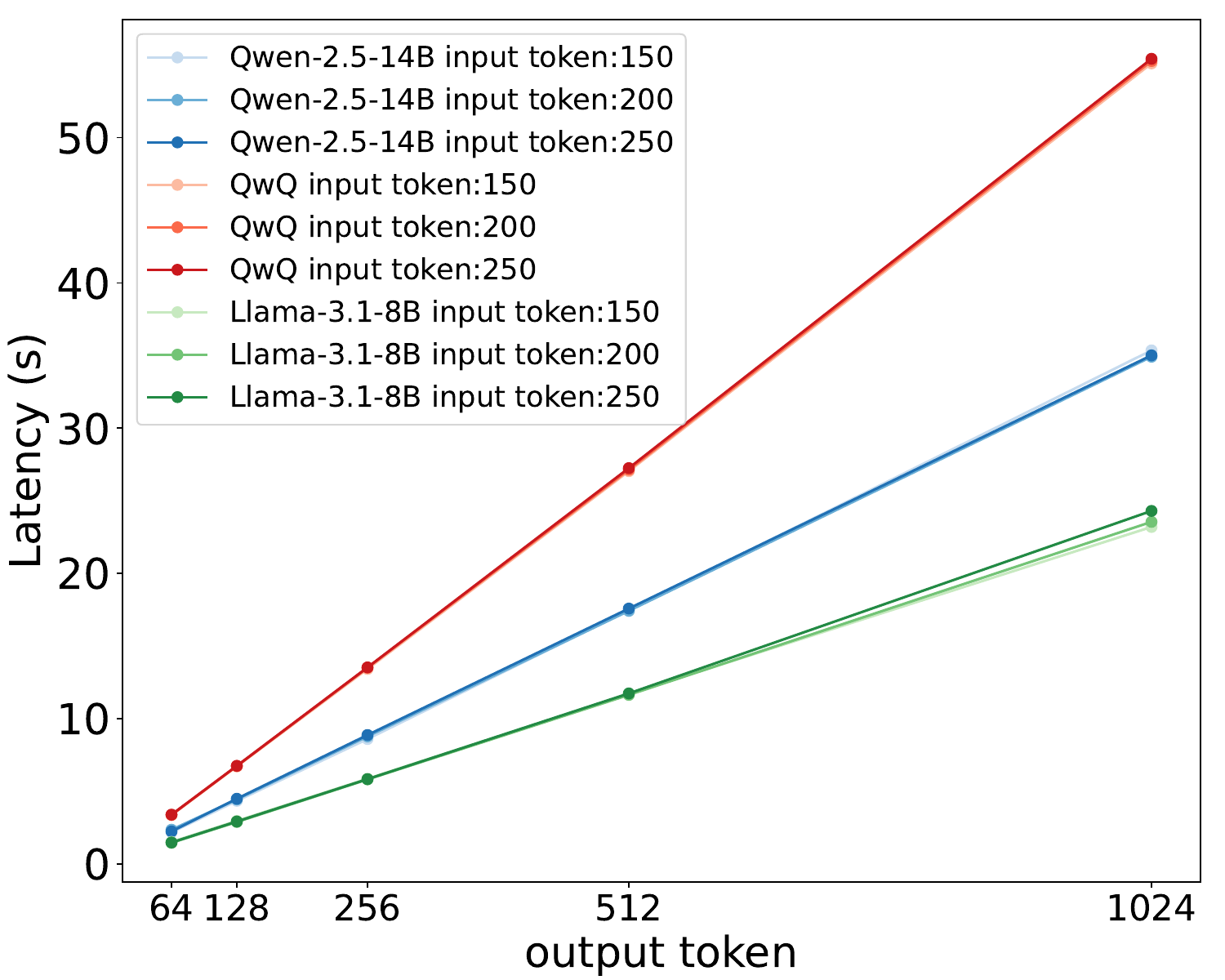}
    \caption{The mapping between output tokens and inference latency. The impact of different input length is quite negligible. And the inference latency is almost linearly correlated to the number of output tokens.}
    \label{fig:token_latency}
\end{figure}

In most cases, the number of input token length falls within the range of 150-250. Therefore, we tested the mapping between output token and inference latency for three models under input token counts: 150, 200, and 250, as shown in Figure \ref{fig:token_latency}.

\begin{figure*}[!htbp]
    \centering
    \includegraphics[width=0.9\textwidth]{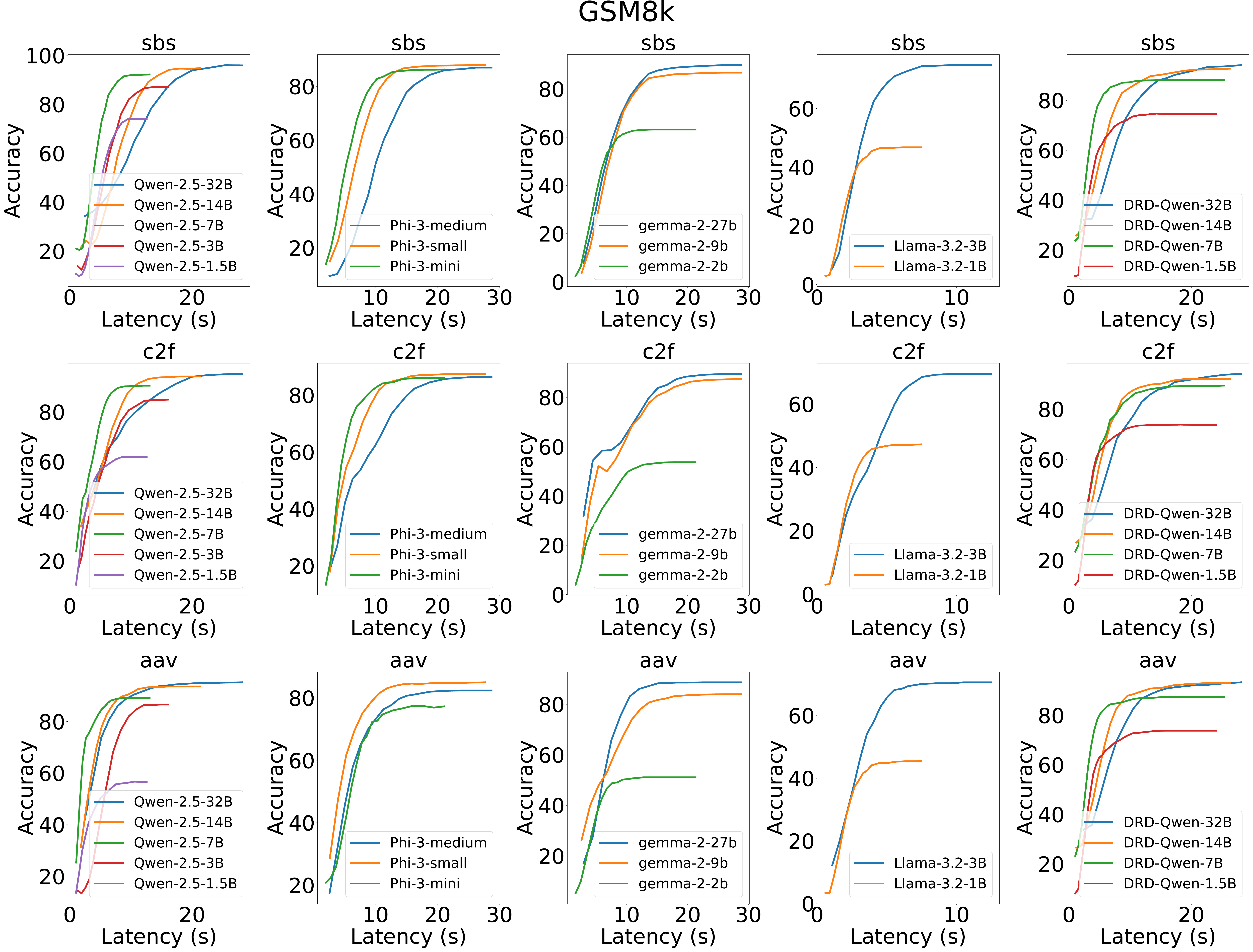}
    \caption{Models' performance under inference latency budget on GSM8K dataset.}
    \label{fig:findings5_gsm8k_all_plot}
    \includegraphics[width=0.9\textwidth]{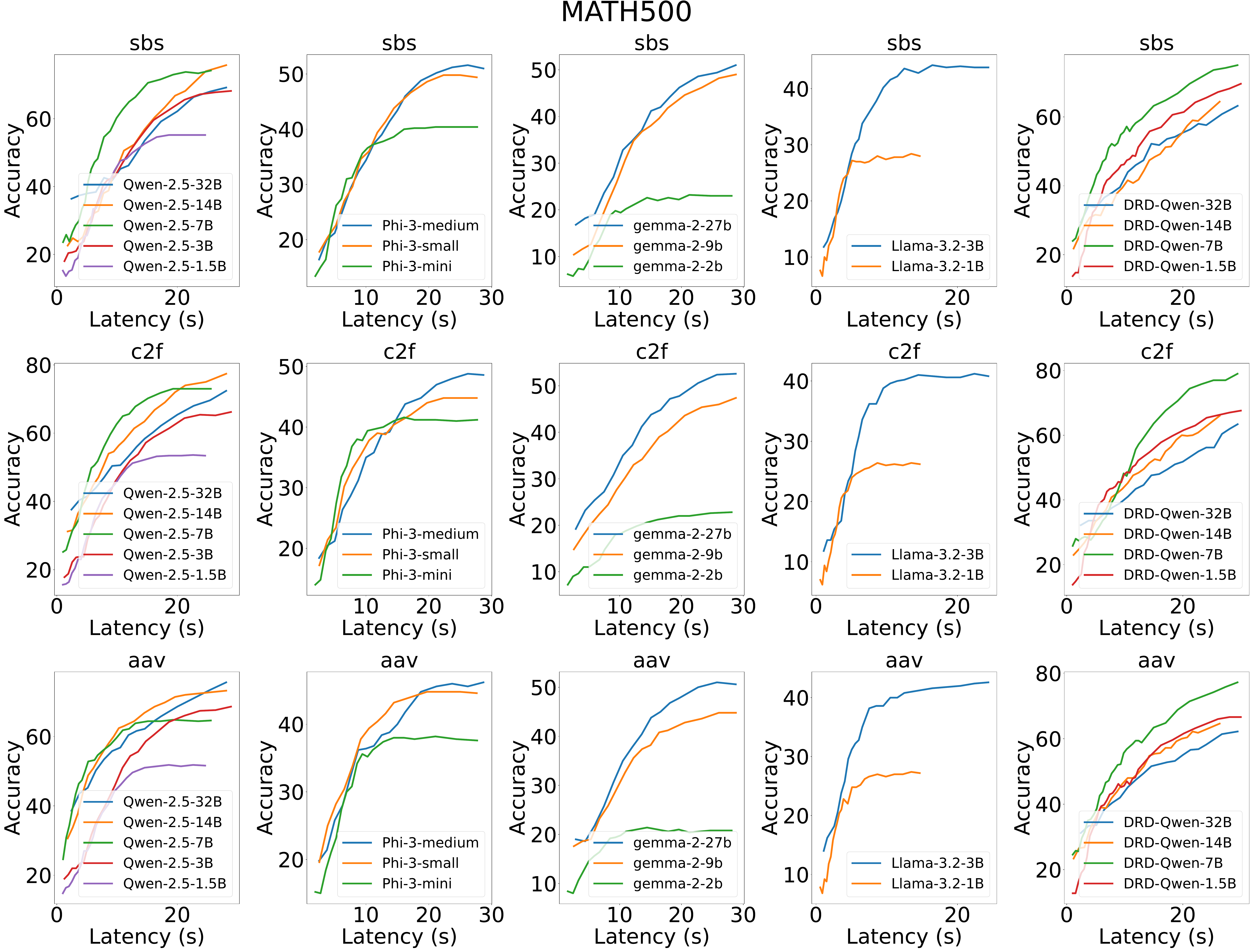}
    \caption{Models' performance under inference latency budget on MATH500 dataset.}
    \label{fig:findings5_math_all_plot}
\end{figure*}

The results indicate that, within output token of 1024, the impact of input length on latency mapping is negligible. Consequently, we applied mapping calculation with input token as 200 in Section \ref{sec:findings 5}. In Figure \ref{fig:findings5_gsm8k_all_plot} and \ref{fig:findings5_math_all_plot}, we show the evaluation of the Qwen-2.5, Phi-3, gemma-2, Llama-3.2, and DRD-Qwen series on the GSM8K and MATH500 datasets, prompted in sbs, c2f, and aav styles.

\end{document}